\documentclass[runningheads]{llncs}

\RequirePackage{rotating}
\PassOptionsToPackage{dvipsnames}{xcolor}

\AtBeginDocument{%
	}
\usepackage[T1]{fontenc}		
\usepackage{amssymb}
\usepackage[english]{babel}
\usepackage[utf8]{inputenc}
\usepackage{amsmath}
\usepackage{hyperref}
\usepackage{booktabs}
\usepackage{mathtools}
\usepackage[nohyperlinks,printonlyused,withpage]{acronym}
\usepackage{tikz}
\usepackage{longtable}
\usepackage{enumitem}
\usepackage{siunitx}
\usepackage{comment}
\usepackage{subfig}
\usepackage{tabularx} 
\usetikzlibrary{decorations,matrix,shapes,arrows,arrows.meta,positioning,calc,fit}
\usepackage{varwidth}
\usepackage[title,toc]{appendix}
\usepackage{graphicx}
\usepackage{multirow}
\usepackage{xcolor, colortbl}
\usepackage{placeins}
\usepackage{slashbox}
\usepackage{wrapfig}
\usepackage{multirow}
\usepackage{floatrow}
\newfloatcommand{capbtabbox}{table}[][\FBwidth]
\usepackage{color}

\tikzset{
	-Latex,auto,node distance =1 cm and 1 cm,semithick,
	state/.style ={ellipse, draw, minimum width = 0.7 cm},
	point/.style = {circle, draw, inner sep=0.04cm,fill,node contents={}},
	bidirected/.style={Latex-Latex,dashed},
	el/.style = {inner sep=2pt, align=left, sloped}
}
\makeatletter
\newcommand{\printfnsymbol}[1]{%
	\textsuperscript{\@fnsymbol{#1}}%
}
\makeatother
\newcommand{\ALL}[1]{\textcolor{magenta}{all: #1}}
\newcommand{\CN}[1]{\textcolor{red}{CN: #1}}
\newcommand{\TK}[1]{\textcolor{blue}{TK: #1}}
\newcommand{\LP}[1]{\textcolor{green}{LP: #1}}
\newcommand{\RG}[1]{\textcolor{teal}{RG: #1}}
\newcommand{\JR}[1]{\textcolor{orange}{JR: #1}}

\renewcommand{\ALL}[1]{}
\renewcommand{\CN}[1]{}
\renewcommand{\TK}[1]{}
\renewcommand{\LP}[1]{}
\renewcommand{\RG}[1]{}
\renewcommand{\JR}[1]{}

\DeclareMathOperator{\BB}{BB}
\DeclareMathOperator{\IRRW}{IRRW}
\DeclareMathOperator{\RRW}{RRW}
\DeclareMathOperator{\ACE}{ACE}
\DeclareMathOperator{\RCE}{RCE}
\DeclareMathOperator{\APE}{APE}
\DeclareMathOperator{\RPE}{RPE}
\DeclareMathOperator{\DoOp}{do}
\newcommand{\Do}[1]{\DoOp({#1})}

\definecolor{lightergray}{gray}{0.85}

\newcolumntype{L}[1]{>{\raggedright\let\newline\\\arraybackslash\hspace{0pt}}m{#1}}
\newcolumntype{C}[1]{>{\centering\let\newline\\\arraybackslash\hspace{0pt}}m{#1}}
\newcolumntype{R}[1]{>{\raggedleft\let\newline\\\arraybackslash\hspace{0pt}}m{#1}}

\hypersetup{draft}

\begin{document}
	\tikzstyle{VertexStyle} = [shape = circle, minimum width = 6ex, 
	\tikzstyle{EdgeStyle}   = [->,>=stealth']

	\title{Causal Bayesian Networks for Data-driven Safety Analysis of Complex Systems}
	
	
	\author{Roman Gansch\inst{1}\thanks{These authors contributed equally to this work.}\and Lina Putze\inst{2}\printfnsymbol{1} \and Tjark Koopmann \inst{2} \and \\ Jan Reich\inst{3} \and Christian Neurohr\inst{2}}

	\institute{Robert Bosch GmbH, Corporate Research, Renningen, Germany\\ \email{roman.gansch@de.bosch.com} \and
	German Aerospace Center (DLR) e.V., Institute of Systems Engineering for Future Mobility, Oldenburg, Germany\\
	\email{\{lina.putze, tjark.koopmann, christian.neurohr\}@dlr.de} \and 
	Fraunhofer Institute for Experimental Software Engineering (IESE), Kaiserslautern, Germany \\ \email{jan.reich@iese.fraunhofer.de} }

	\authorrunning{R. Gansch et al.}
	\titlerunning{Causal Bayesian Networks for Safety Analysis}
	
	\maketitle
	\begin{abstract}
		Ensuring safe operation of safety-critical complex systems interacting with their environment poses significant challenges, particularly when the system's world model relies on machine learning algorithms to process the perception input.
		A comprehensive safety argumentation requires knowledge of how faults or functional insufficiencies propagate through the system and interact with external factors, to manage their safety impact.
		While statistical analysis approaches can support the safety assessment, associative reasoning alone is neither sufficient for the safety argumentation nor for the identification and investigation of safety measures.		
		A causal understanding of the system and its interaction with the environment is crucial for safeguarding safety-critical complex systems. It allows to transfer and generalize knowledge, such as insights gained from testing, and facilitates the identification of potential improvements.
		This work explores using causal Bayesian networks to model the system's causalities for safety analysis, and proposes measures to assess causal influences based on Pearl's framework of causal inference.
		We compare the approach of causal Bayesian networks to the well-established fault tree analysis, outlining advantages and limitations. 
		In particular, we examine importance metrics typically employed in fault tree analysis as foundation to discuss suitable causal metrics.
		An evaluation is performed on the example of a perception system for automated driving.
		Overall, this work presents an approach for causal reasoning in safety analysis that enables the integration of data-driven and expert-based knowledge to account for uncertainties arising from complex systems operating in open environments.
	\end{abstract}

	\section{Introduction}
	\label{sec:introduction}

Ensuring the safe operation of safety-critical complex systems that interact with their environment based on information obtained by perception components is a challenging endeavor. In particular, such perception components often rely on complex algorithms like machine learning to construct a world model out of sensory input. The verification and validation of these is notoriously difficult and reliance on statistical, non-causal metrics is unsatisfactory from a safety perspective \cite{damm2018perspectives}. 
Essentially, safety engineers are not interested in associations, but in causal explanations of how faults and failures are propagated within a system. An example for this is the well-known fault-error-failure model of Avizienis et al.\ \cite{avizienis2004basic}.
Therefore, it is indispensable to integrate causal metrics for the safeguarding of safety-critical systems, especially regarding the perception components.
In order to obtain causal information about complex systems and faults in their perception, Kramer et al.\ suggest to adapt fault tree analysis (FTA) for this task \cite{kramer2020identification}. However, restricting the causal graph structure to trees drastically limits modeling possibilities. Moreover, the quantification of fault trees rests on the assumptions of stochastic independence of its base events which can conflict with handling of confounders. To overcome these inadequacies, we propose to use causal Bayesian networks (CBNs) to model and analyze the causalities behind fault propagation in complex systems, based on Pearl's causal theory \cite{pearl2009causality}. 

The contributions of this work can be summarized as follows:
\begin{itemize}
	\item[$\bullet$] a novel approach relying on CBNs combined with suitable causal metrics,
	\item[$\bullet$] a comparison between fault trees and CBNs, with a focus on quantification,
	\item[$\bullet$] evaluation of the approach for an automated driving perception system. 
\end{itemize}

Following the introduction of \autoref{sec:introduction}, we cover the preliminaries and related work in \autoref{sec:preliminaries}, i.e.\ the role of causality in safeguarding complex systems. Section \ref{sec:example} covers in detail the example of an perception system for automated driving, before concluding with \autoref{sec:conclusion}.
	
	\section{Causality in Safety Analysis}
	\label{sec:preliminaries}

Safety of technical systems is achieved by applying multiple measures in combination during the complete system life-cycle.
An integral part of safety engineering is the safety analysis. This analysis supplements the synthesis step during design and verifies that certain design criteria are fulfilled. 
Goals of the safety analysis are to identify faults and functional insufficiencies that propagate through the system and lead to hazards, as well as estimating the overall residual risk.
A common approach is to model the fault propagation pathways (fault-error-failure chain) \cite{avizienis2004basic}. A wide range of methods have been adopted in industrial practice, each of which is useful within a certain context.
Since the advent of highly automated systems adaption of established analysis methods have been done. For example, the ISO 21448 standard \cite{iso21448} recommends the application of \emph{System-Theoric Process Analysis} (STPA) and \emph{Cause Tree Analysis} (CTA) (an adaption of FTA) in combination with statistical analysis of the occurrence of triggering conditions (TCs). However, these simple methods often fail to include the complex relations or neglect the causal mechanisms. In particular, when artificial intelligence is utilized these practices are either too abstract or rely on unsatisfiable assumptions.
In this paper we explore the use of CBNs focusing on a comparison with FTA. CBNs offer a quantitative approach to investigate causal influences on safety based on a data-driven approach. In contrast to FTA this approach does not require independence of specific factors and allows to model complex dependencies. Further it enables a shift from deterministic causation to probabilistic causation, i.e., a cause does not always lead to an effect, but rather might be suppressed due to factors not included in the model.
In the next section we provide an overview of causal inference with examples illustrating its relevance for safety engineering.

\FloatBarrier
\subsection{Causal Inference}

Causal theory formally describes the influence of a cause on an effect. It has been pioneered and frequently applied in the field of economics, sociology and medicine \cite{pearl2009causality,peters2017}. Recently, causal inference also has gained a lot of traction in the engineering domain \cite{nyberg2013failure,issa2022use,niu2023causal,maier2024testing,koopmann2025grasping}.
Pearl describes causality with a 3-step ladder: 
The first step \emph{association} is about predicting the outcome $Y$ under observations $X$ which can be described by purely statistical quantities.
The second step \emph{intervention} provides an answer to causal queries of the form: "Which effect $Y$ can be observed in a population if the value of $X$ is intervened on?". 
The third step \emph{counterfactual} is highest form of causality. It answers the questions "What would have been $Y$ if $X$ had been intervened on?".

Causal intervention queries, and even some counterfactual queries, can be answered by means of intervention experiments or by estimation from observational data.
In an intervention experiment the intervention is performed while collecting data, like e.g. in randomized control trials (RCTs). 
However, performing an intervention experiment is often not possible due to constraints in the experimental environment or ethical considerations. \\
The do-calculus introduced by J. Pearl allows to evaluate interventional queries from observational data without additional experiments, a characteristic that is termed identifiability \cite{pearl2009causality}.
The do-calculus provides inference rules to reformulate an interventional query in the form $P(Y|\Do{X=x})$ to an expression only containing conditional probabilities obtainable from observational data.
In order to apply the do-calculus, a causal model expressing the cause-effect relationships between variables is required. In this paper we use graphical causal models, which are directed acyclic graphs (DAG) to model these. In this notation, an intervention $\Do{x}$ can be seen as removing all incoming edges to $X$.

\FloatBarrier
\subsection{Causal Bayesian Networks}
The advantage of using a graphical notation for the causal models is, that it integrates well with quantification of the variables as it is inherently similar to Bayesian networks (BN) \cite{fenton2018risk} since both are built from DAGs.
In a BN the direction of the arrows indicate the order of factorization of the joint probability distribution into conditional probability tables (CPT).
The order of factorization can be freely chosen as it is only based on correlation which can be reformulated by the Bayes theorem.
By selecting the arrow directions according to the causal relationships we obtain a causal Bayesian network (CBN).\\
In a CBN correlational as well as causal inferences can be performed.
Previous work has explored the use of BN and CBN for safety, cf.~\autoref{tab:related_work}.
We distinguish between BNs that use only correlational structures and CBNs that use  causal structures. 
A BN can only be used for correlational inference, while a CBN has the advantage of a causal structure interpretation for the modeler.

\begin{wraptable}[9]{R}{0.47\textwidth}
	\small
	\centering
	\vspace{-0.7cm}
	\begin{tabular}{lcc}
		\toprule
		& \multicolumn{2}{c}{Inference}  \\
		Structure & Correlation & Causation \\
		\midrule 
		Correlation & \cite{fenton2018risk,jesenski2019generation,qiu2021parameter} & - \\
		Causation & \cite{adee2021systematic,adee2021discovery,thomas2023toward,werling2025safety} &
		\cite{deletang2021causal,maier2024testing,nyberg2013failure,koopmann2025grasping,jiang2024generation}\\
		\bottomrule
	\end{tabular}
	\caption{Related work on (causal) Bayesian network for safety analysis.}
	\label{tab:related_work}
\end{wraptable}

Building a CBN model can be separated into the task of structuring the DAG and quantifying the CPTs.
For both either an expert-based or data-driven approach can be chosen. 
Learning the causal structure from data is referred to as \emph{structure learning} or \emph{causal discovery} \cite{peters2017}. 
Learning the conditional distribution from data is termed \emph{parameter learning}. 
The graph structure and the parameters can also be obtained through \emph{expert knowledge} \cite{neurohr2021criticality} or by combined approaches feeding expert-based constraints into learning algorithms.
For our proposed approach of using CBNs for safety analysis, we favor the expert-based approach to define the structure and parameter learning from data as it combines the best of both worlds.
An expert-based structure is more appropriate to argue to capture the underlying causality, while expert judgment on quantifying probabilities is susceptible to bias \cite{skjong2001expert}.

The nodes in the CBN correspond to random variables whose value ranges can be dichotomous, categorical, ranked, or even continuous. 
Dichotomous variables only contain two states which have a binary true/false character.
A FTA model only consists of these kind of variables. 
For a SOTIF oriented safety analysis the triggering conditions in the domain have to be included. 
These often require a continuous distribution or a mapping to categorical variables with multiple states (e.g., weather: sun, cloudy, rain, snow). 
While inference calculations can generally be performed on continuous multivariate distributions, it requires significant computational resources.
Further, accurately quantifying continuous distributions demands a large amount of data. 
In practice, continuous distributions are either discretized to categorical nodes or described as parametric distributions that allow to analytically pre-solve the necessary integrals. 
The CBN examples in this paper only use categorical variables as these are similar to the dichotomous variables used in FTA. For implementation we use the python library pyAgrum \cite{ducamp2020agrum}.

\FloatBarrier
\subsection{From Correlation to Causation}
\label{subsec:CorrToCaus}

To grasp the differences between association and causality and how it impacts safety engineering we examine some examples. 

\captionsetup[subfloat]{justification=centering}
\begin{figure}[htb!]
	\captionsetup[subfigure]{font=footnotesize}
		\subfloat[][\hspace{-0.1cm}CBN]{\includegraphics[width=0.2\textwidth]{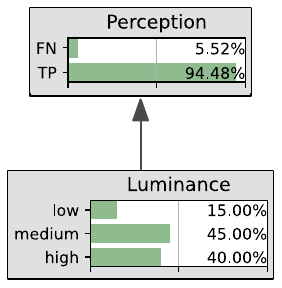}\label{fig:CBN2nodes_CBN}}
		\vline
		\subfloat[][\hspace{-0.1cm}$P(P|L)$]{\includegraphics[width=0.2\textwidth]{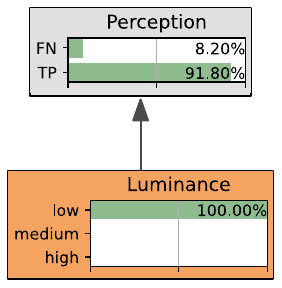}\label{fig:CBN2nodes_Cor1}}
		\subfloat[][\hspace{-0.1cm}$P(L|P)$]{\includegraphics[width=0.2\textwidth]{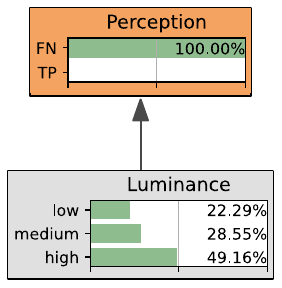}\label{fig:CBN2nodes_Cor2}}
		\vline
		\subfloat[][\hspace{-0.12cm}$P(P|do(low))$]{\includegraphics[width=0.2\textwidth]{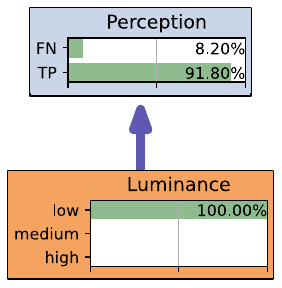}\label{fig:CBN2nodes_Caus1}}
		\subfloat[][\hspace{-0.1cm}$P(L|do(FN))$]{\includegraphics[width=0.2\textwidth]{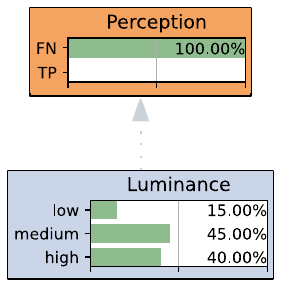}\label{fig:CBN2nodes_Caus2}}
	\caption{Causal Bayesian network (CBN) consisting of two nodes: Perception (P) and Luminance (L) (left). Correlational inference is agnostic to the causal direction (center). Causal inference depends on the causal direction (right). Bold blue indicates a causal inference along a causal pathway, while a dashed gray indicates deletion due to intervention.}
	\label{fig:CBN2nodes}
\end{figure}

First, consider a simple two node graph as shown in Figure~\ref{fig:CBN2nodes_CBN}. 
It represents the causal mechanism of a typical perception example for automated driving, where we are interested in the perception performance under the influence of a triggering condition. 
The upper node Perception (P) represents the performance of a camera-based object detection in terms of false negatives (FNs) and true positives (TPs). 
The lower node Luminance (L) corresponds to the light intensity of an object, ranked from low to high. 
To a human it is intuitively clear that luminance affects the performance of the camera-based object detection and not vice-versa.
However, based on association alone we cannot distinguish both causal directions, cf.~Figure~\ref{fig:CBN2nodes_Cor1}~and~\ref{fig:CBN2nodes_Cor2}. 
Conditioning on either of both variables leads to changes in the distribution of the outcome variable compared to the observed distribution of Figure~\ref{fig:CBN2nodes_CBN}. 
The correlation between the two variables is agnostic to the underlying cause-effect structure.
In contrast, causal intervention queries can expose the cause-effect structure.  
Intervening on luminance has an effect on the perception, while changing the perception result does not affect the luminance, cf.~Figure~\ref{fig:CBN2nodes_Caus1} and \ref{fig:CBN2nodes_Caus2}.
Whether an intervention reveals some effect depends on the direction of the causal paths.

\FloatBarrier

\begin{figure}[htb!]
	\vspace{-0.5cm}
	\centering
	\subfloat[][]{\includegraphics[width=0.19\textwidth]{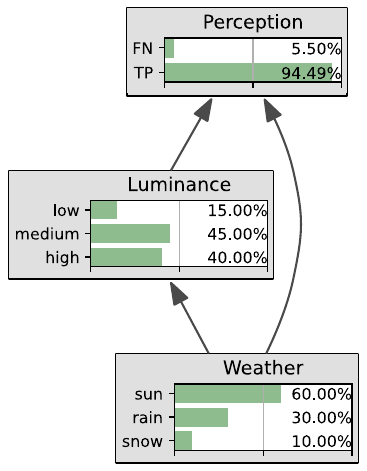}}\label{fig:conf_CBN}
	\subfloat[][]{\includegraphics[width=0.19\textwidth]{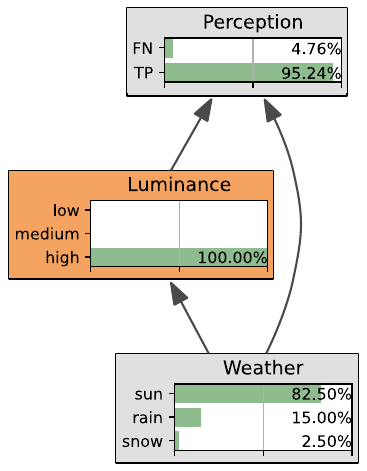}}\label{fig:conf_Cor}
	\subfloat[][]{\includegraphics[width=0.19\textwidth]{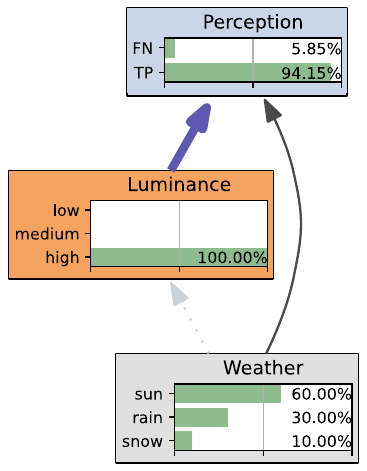}}\label{fig:conf_Caus}
	\vline
	\hspace{0.08cm}
	\subfloat[][]{\includegraphics[width=0.16\textwidth]{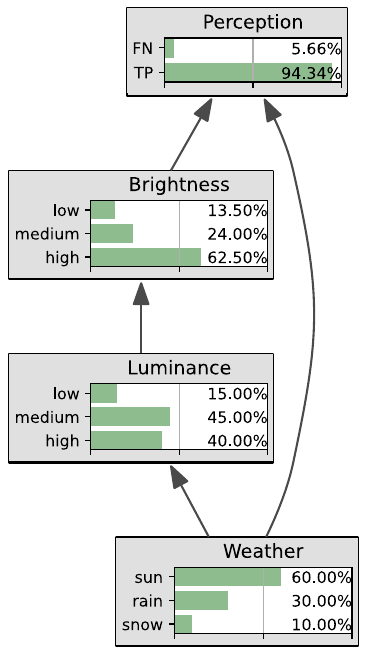}}\quad
	\subfloat[][]{\includegraphics[width=0.16\textwidth]{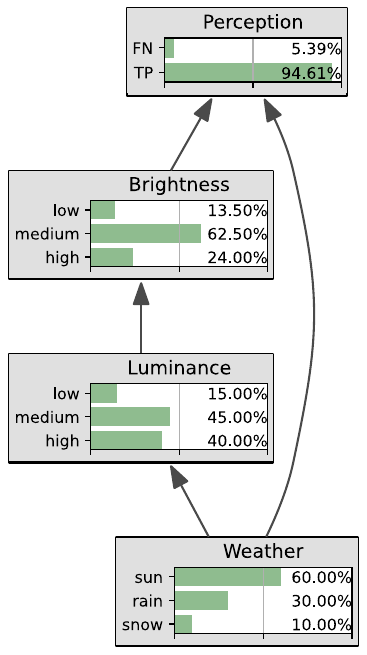}}
	\caption{(a) Causal Bayesian network for perception performance influenced by luminance and weather. Analysis results based on (b) correlation with  $P(P|L=high)$ and (c) causation with  $P(P|\Do{L=high})$. Safety measure design based on (d) correlation and (e) causation. Probabilities are given in \autoref{tab:conditional_probabilities_confounding}.}
	\label{fig:confounding_example}
\end{figure}
Another distinction between correlational and causal queries arises due to so-called confounding.
The issue of confounding is encountered when there exists a common-cause, like Weather in \autoref{fig:confounding_example}. 
From the result of the correlational query $P(P|L)$ it seems that a high luminance  improves the perception performance, cf.~ \autoref{fig:confounding_example}(b). 
But this result is affected by the change in the  distribution of the weather conditions when conditioning on luminance.
If we investigate the causal effect based on an intervention, i.e. if we keep the observed distribution of the weather conditions, we encounter indeed that high luminance on its own will decrease the performance of the perception, cf.~ \autoref{fig:confounding_example}(c). 

The presented results have implications for the design of potential safety measures.
In the given example, this can be a simple mechanism that modifies the brightness of the camera pictures in the pre-processing step of an AI-based object detection. 
Based on the result of the correlation analysis, a safety engineer will favor high brightness as a higher luminance correlates with better performance, leading e.g. to the CBN of \autoref{fig:confounding_example}(d). 
Compared to the marginal FN rate of the unmodified structure, the FN rate including the safety measure actually deteriorates from $5.5\%$ to $5.66\%$. 
This demonstrates how interpreting correlation as causation can lead to a counterproductive system design.
In contrast, applying the results of the intervention analysis to design safety measures, a shift of towards medium brightness seems most beneficial, resulting in the CBN of \autoref{fig:confounding_example}(e). 
Here, the marginal FN rate has actually improved from $5.5\%$ to $5.39\%$ providing an increased performance. 
	
	\section{Use Case: Perception of Automated Driving Systems}
	\label{sec:example}

\begin{wrapfigure}[11]{R}{0.45\textwidth}
	\vspace{-0.8cm}
	\centering
	\includegraphics[width=\textwidth]{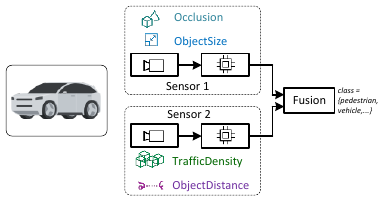}
	\caption{Example architecture for an ADS perception use case.}
	\label{fig:use_case_architecture}
\end{wrapfigure}

\FloatBarrier
To illustrate the application of CBNs and causal importance metrics for safety analysis of complex systems and to compare them to a classical FTA, we consider as example a perception subsystem commonly used for ADSs, cf.~\autoref{fig:use_case_architecture}. 
Although the data is not from an actual implemented perception system, it closely reflects a potential real-world application.
The perception subsystem consists of two redundant sensor modalities each with a software-based perception algorithm to classify objects from sensor data. 
Both modalities may employ a different sensing principles and different perception algorithms each with specific functional insufficiencies and corresponding sensitivities to environmental TCs, e.g., Occlusion/ObjectSize for Sensor 1 and TrafficDensity/ObjectDistance for Sensor~2. 
The performance reduction of  each sensor as well as the perception subsystem can be captured using the FN rate as indicator.

\subsection{Causal Modeling}

\begin{figure*}[htb!]
	\centering
	\subfloat[][]{\includegraphics[width=0.42\textwidth]{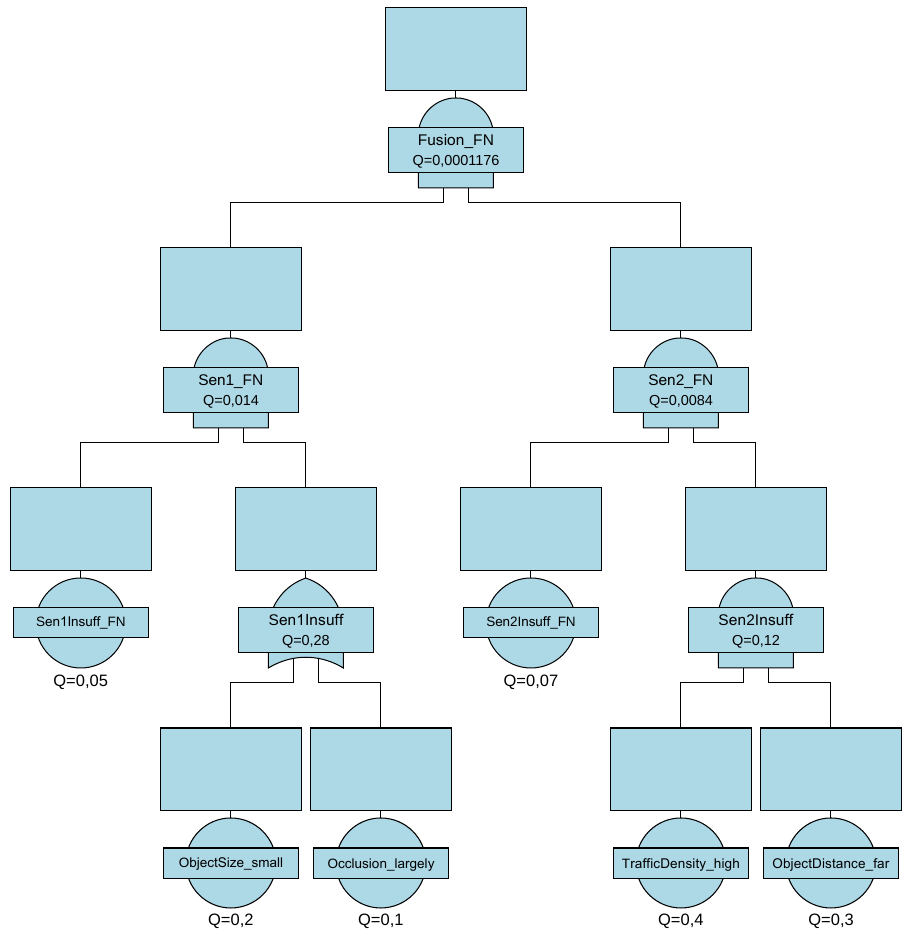}\label{fig:UC_modelFTA}}
	\quad
	\subfloat[][]{\includegraphics[width=0.48\textwidth]{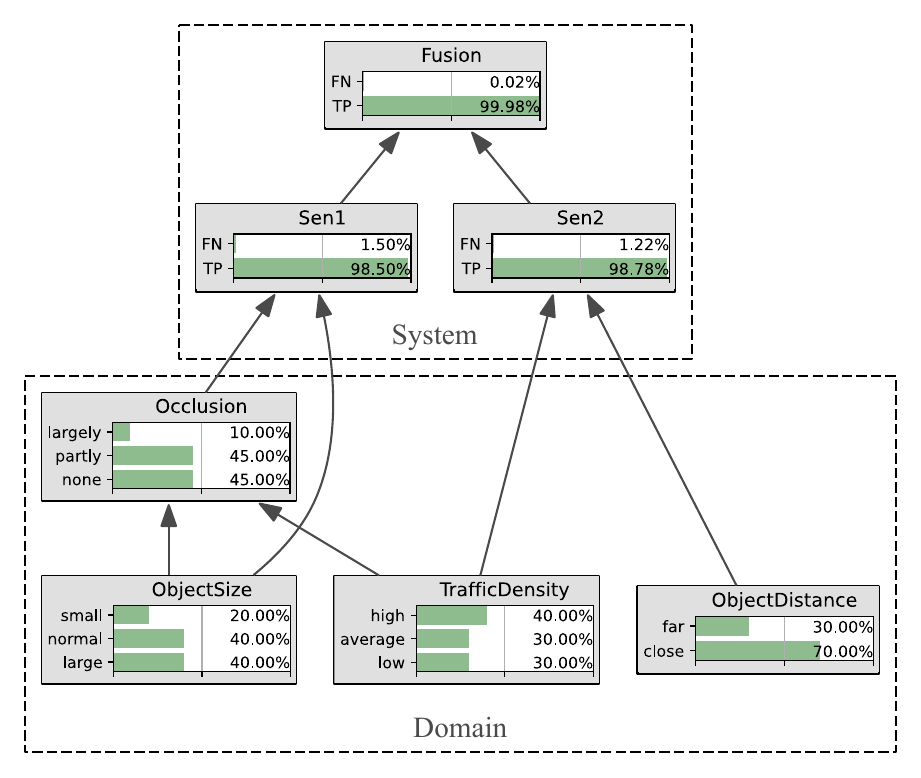}\label{fig:UC_modelCBN}}
	\caption{Perception example modeled with (a) FTA and (b) CBN. The corresponding (conditional) probabilities are given in Table \ref{tab:conditional_probabilities_use_case_domain_nodes} and \ref{tab:conditional_probabilities_use_case_system_nodes}.}
	\label{fig:use_case_models}
\end{figure*}

A safety analysis seeks to identify the causal pathways of faults and functional insufficiencies emerging into system failures and pinpoint areas of improvement. 
A straightforward approach to model the perception system of \autoref{fig:use_case_architecture} in a FTA as proposed for SOTIF oriented analysis \cite{zeller2022component} is shown in Figure \ref{fig:UC_modelFTA}. 
The TCs are included as base events that activate a sensor insufficiency. 
As required by FTA, the base events are assumed to be independent.

Figure \ref{fig:UC_modelCBN} models the same example as CBN. 
In contrast to FTA, CBNs are not restricted to a tree structure with independent base events. 
While such tree structure is usually adequate for modeling dependencies of a well-defined system architecture, domain-level nodes often exhibit complex interdependencies, necessitating a less restrictive framework.
By modeling the example as CBN, dependencies of Occlusion on TrafficDensity and ObjectSize can be taken into account.
Further, in the CBN the nodes representing active insufficiencies (Sen1Insuff, Sen2Insuff) are removed. 
These nodes do not represent actual causal artifacts but rather serve as subsidiary constructs to represent probabilistic relations in the FTA, which can be directly integrated into the CPTs of Sen1 and Sen2.

Without the restriction to dichotomous nodes imposed by FTA, the nodes of a CBN can be discretized into categorical variables. 
For example, Occlusion can be expanded into \emph{largely, partly} and \emph{none}. 
The refinement of node values is a valuable tool to approximate reality more closely.
To enable a comparison, the CPTs assigned to the nodes in this example preserve the marginal rates of the FTA base events. 
However, in the domain part the additional relations between nodes are reflected in the CPTs, cf.~\autoref{tab:conditional_probabilities_use_case_domain_nodes}, and in the system the AND/OR gates have been modified to a non-perfect relation, deviating a few percent, cf.~\autoref{tab:conditional_probabilities_use_case_system_nodes}. 
This reflects the semantic abstraction as we do not model on a detailed level of bits and pixels but rather on a higher abstraction level of objects in a camera picture. 
Therefore, we can not model fully deterministic fault propagation and have to account for some error terms due to the abstraction.

\subsection{Causal Safety Metrics}
\label{subsec:safety_metric}
CBNs as well as fault trees allow for quantitative evaluation of fault and failure propagation through the system.
The state of the art in FTA are importance metrics that assess the impact of base events
$(N_i)_{i\in I}$ on the top level event ($T$) to provide a ranking. 
Several importance metrics have been defined in literature, each providing a different ranking order \cite{birnbaum1968importance,espiritu2007component,ruijters2015}. 
For comparison with causal analysis we focus on the Birnbaum ($\BB$) importance  and the Risk Reduction Worth ($\RRW$):
{\footnotesize
\begin{equation*}
\BB=\frac{\partial P(T=fail)}{\partial P(N_i=fail)}, \qquad  \RRW=\frac{P(T=fail)}{P(T=fail|N_i=\neg fail)}.
\end{equation*}}

The $\BB$ importance provides a sensitivity metric for a top event  failure to a base event. 
For independent basic events it can also be written as $\BB=P(T=fail|N_i=fail)-P(T=fail|N_i=\neg fail)$. 
It is also referred to as structural importance since it only responds to structural changes of the fault tree and not to the failure rates of the basic event.
In contrast, the $\RRW$  measures the potential reduction in the probability of the top level event if the base event does not occur. 
\begin{wraptable}[13]{R}{0.45\textwidth}
	\vspace{-0.5cm}
	\centering
	\footnotesize
	\begin{tabular}{L{2.2cm}cccc}
		\toprule
		& \multicolumn{2}{c}{$\BB$ \tiny{$\left(1\mathrm{e}{-4}\right)$}} & \multicolumn{2}{c}{$\RRW$}\\
		\cmidrule{2-3} \cmidrule{4-5}
		Triggering Condition & FTA & CBN & FTA & CBN \\
		\midrule
		ObjectSize & $3.78$ & $3.12$ & $2.8$ & $1.50$\\
		Occlusion & $3.36$ & $4.39$ & $1.4$ & $1.33$\\ 
		TrafficDensity & $2.94$ & $3.35$ & $\infty$ & $3.59$\\
		ObjectDistance & $3.92$ & $3.52$ & $\infty$ & $2.31$\\
		\bottomrule
	\end{tabular}
	\caption{$\BB$ and $\RRW$ importance for the TCs in the FTA and CBN model, respectively.}
	\label{tbl:importance}
\end{wraptable}

Table \ref{tbl:importance} provides the calculated importances for both, the fault tree as well as the CBN. 
The partial derivative  of the $\BB$ importance is calculated by setting small soft evidences on the nodes (about $1\%$) and estimating the difference quotient. 
We observe slight deviations in the results of the FTA and the CBN, which can be explained by the couplings between the TCs and the non-perfect OR/AND gates in the CBN.
A significant difference occurs for the $\BB$ importance of Occlusion, due to the confounding effect of  TrafficDensity.

\begin{figure}[htb!]
	\centering
	\subfloat[][]{\includegraphics[width=0.33\textwidth]{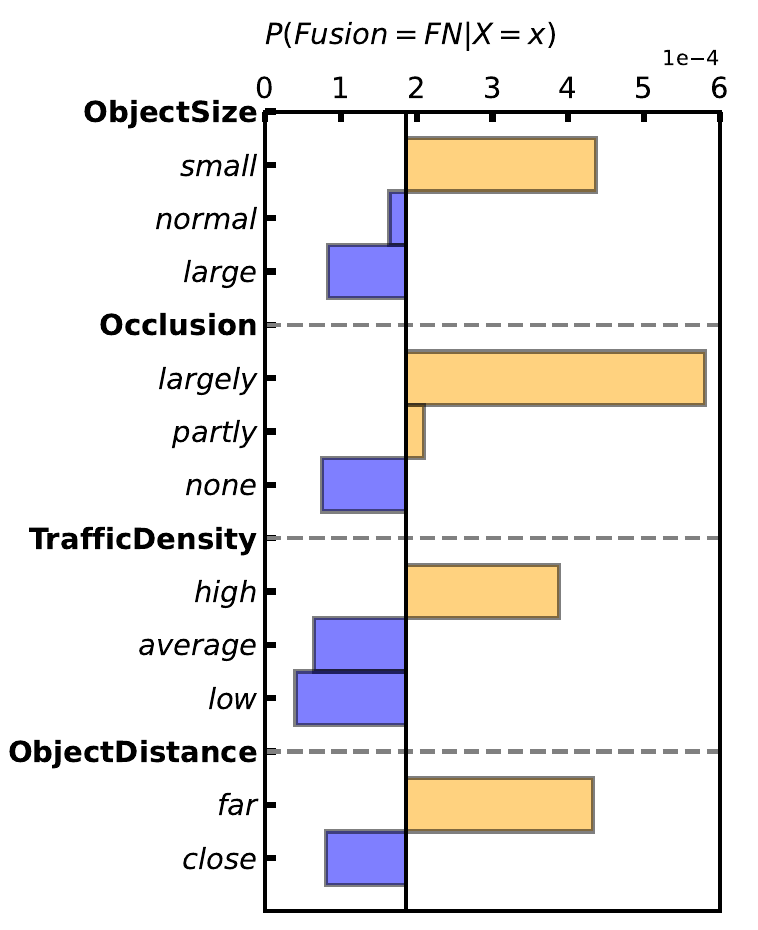}\label{fig:tornadochart_cor}}
	\subfloat[][]{\includegraphics[width=0.33\textwidth]{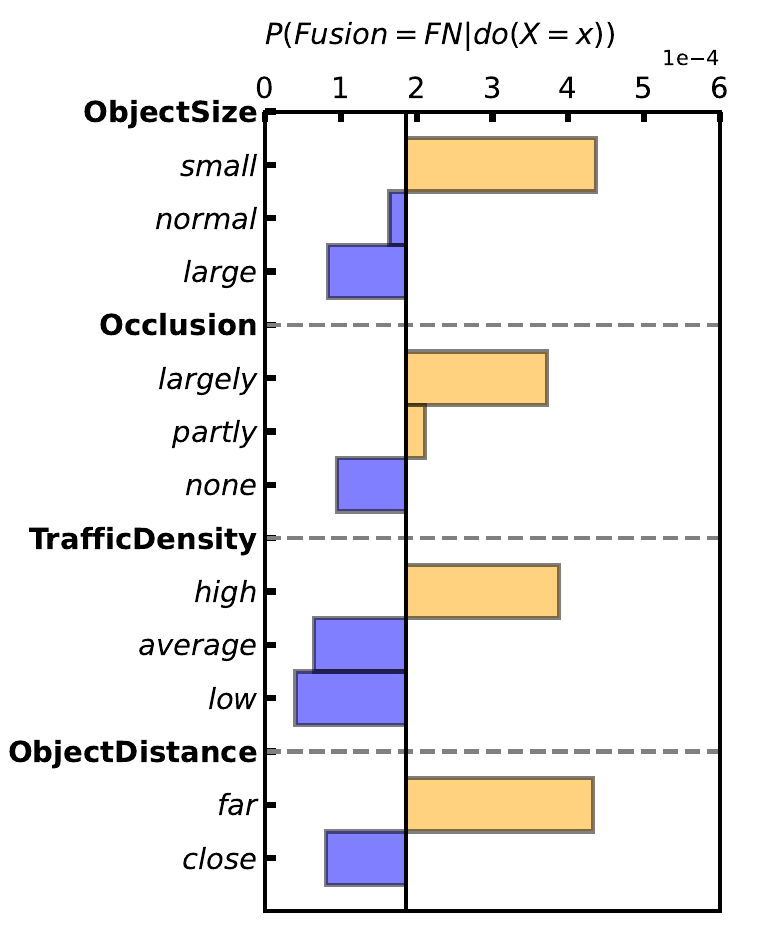}\label{fig:tornadochart_caus}}
	\subfloat[][]{\includegraphics[width=0.33\textwidth]{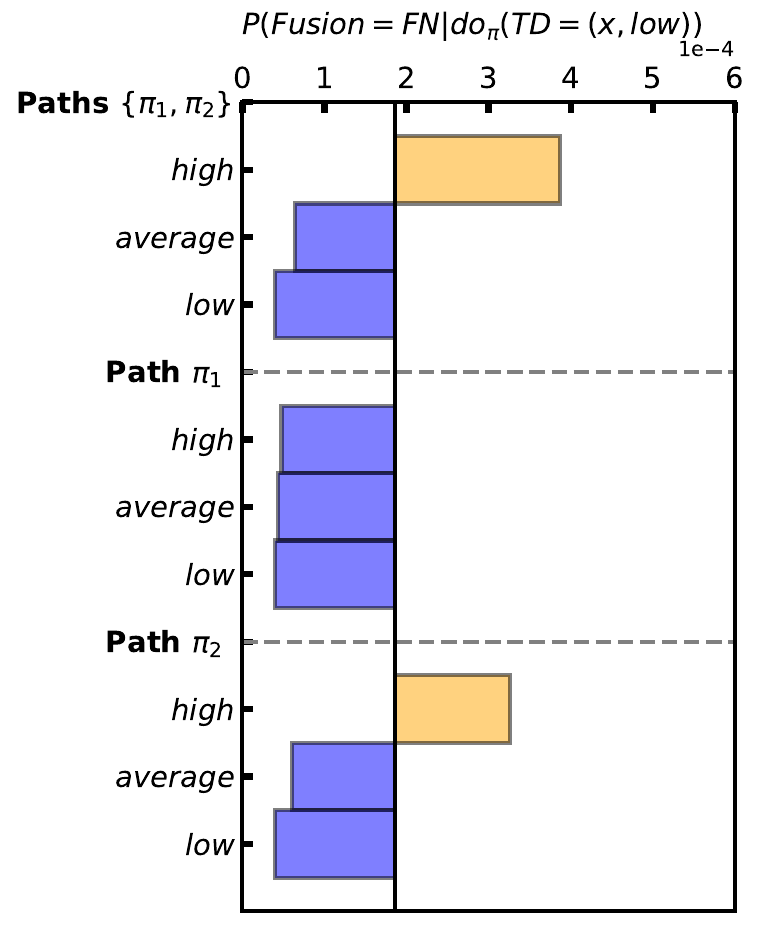}\label{fig:tornadochart_path}}
	\caption{Tornado charts for (a) correlational analysis, (b) causal intervention analysis for all TCs ($X$), and (c) for the categorical analysis of the path-specific effects of TrafficDensity (TD) on Fusion=FN via  $\lbrace\pi_1, \pi_2\rbrace$, $\pi_1$ and $\pi_2$. The vertical line indicates the marginal probability $P(Fusion=FN)$.}
	\label{fig:tornadochart}
\end{figure}

While FTA importance metrics can  be applied to CBNs, caution is required when interpreting the results.
By restricting fault trees to a tree structure with independent base events, confounding effects are eliminated.
This leads to equality of conditional probabilities $P(Y|X=x)$ and interventional probabilities $P(Y|\Do{X=x})$ querying alongside the causal direction of the fault tree.
Consequently,  associative importance metrics can be interpreted causally in the FTA.
However, for structures like the CBNs that cover more complex dependencies, this equality does not apply, as outlined in \autoref{subsec:CorrToCaus}.
\autoref{fig:tornadochart} provides a visual comparison of the conditional and interventional probabilities resulting from the CBN in a tornado chart.
For Occlusion, which has ObjectSize and TrafficDensity as confounding nodes, a significant deviation between correlation and causation can be observed.
This illustrates, the importance of causal metrics for a comprehensive evaluation of CBNs.


In causal literature the average causal effect ($\ACE$) and relative causal effect ($\RCE$) are commonly used \cite{peters2017,koopmann2025grasping}.
These metrics evaluate the  structural importance of a node, similar to the $\BB$ importance.
Both, the $\ACE$ and $\RCE$ are originally defined for dichotomic states as the absolute and relative difference of both possible interventions states.
To apply both metrics for the SOTIF analysis, the metrics need to be generalized to categorical variables.
For TCs it is usually possible to define a reference state $x_{ref}$ representing the nominal conditions to which others are compared, like 'none' for Occlusion.
Thus, we define
{\footnotesize
\begin{equation}
	\label{eq:ace_rce}
	\ACE  =P(Y|\Do{X=x})-P(Y|\Do{X=x_{ref}}),\quad
	\RCE  =\frac{P(Y|\Do{X=x})}{P(Y|\Do{X=x_{ref}})}
\end{equation}}
where the comparative value $x_{ref}$ can either be $\neg x$ for dichotomic analysis or a reference value for a categorical analysis.
For further analysis we consider the $\RCE$ as relative metrics are easier to interpret. 
Safety measures to improve the system have to focus on mitigating the influence of TCs with a high $\RCE$. 
However, similar to BB importance, the $\RCE$ does not consider the overall occurrence of a triggering condition and, hence, may not provide the best improvement of the system performance.
It rather provides an argument to mitigate systematic issues leading to an increase of risk.
To account for the occurrence of the TCs and evaluate how  probability shifts affect the overall system performance, the $\RRW$ can be generalized to categorical variables and transferred to a causal metric, referred to as \emph{Interventional Risk Reduction Worth} ($\IRRW$):
{\footnotesize
\begin{equation*}
	\RRW=\frac{P(Y)}{P(Y|X=x_{ref})}, \qquad
	\IRRW=\frac{P(Y)}{P(Y|\Do{X=x_{ref}})}.
\end{equation*}}

\autoref{tbl:acerce_single} shows the results of the categorical and dichotomic calculation  of $\RCE$, $\RRW$ and $\IRRW$.

\begin{table}[htb!]
	\centering
	\footnotesize
	\begin{tabular}{p{1.5cm}lcccccc}
		\toprule
		\multirow{2}{1cm}{Triggering Condition}& \multirow{2}{*}{State} &\multicolumn{3}{c}{Categorical}&\multicolumn{3}{c}{Dichotomic}\\
		\cmidrule(lr){3-5} \cmidrule(lr){6-8}
		 && $\RCE$ & $\RRW$ & $\IRRW$ & $\RCE$ & $\RRW$ & $\IRRW$ \\
		\midrule
		\multirow{3}{1.4cm}{Object Size}& small & $2.66$ & \cellcolor{lightergray} & \cellcolor{lightergray} & 3.50 & 1.50 & 1.50 \\
		& \cellcolor{lightergray}normal & \cellcolor{lightergray}$1.00$ & \cellcolor{lightergray} & \cellcolor{lightergray} & 0.79 & 0.89 & 0.89 \\
		& large & $0.51$ & \cellcolor{lightergray}\multirow{-3}{*}{$1.14$} & \cellcolor{lightergray}\multirow{-3}{*}{$1.14$} & 0.33 & 0.73& 0.73 \\
		\midrule
		\multirow{3}{1.4cm}{Occlusion} & largely & $3.95$ & \cellcolor{lightergray} & \cellcolor{lightergray} & 2.41 & 1.33 & 1.20 \\
		& partly  & $2.23$ &  \cellcolor{lightergray} & \cellcolor{lightergray} & 1.43 & 1.25 & 1.26 \\
		& \cellcolor{lightergray}none & \cellcolor{lightergray}$1.00$ & \cellcolor{lightergray}\multirow{-3}{*}{$2.49$} & \cellcolor{lightergray}\multirow{-3}{*}{$1.97$} & 0.40 & 0.69 & 0.79 \\ 
		\midrule
		\multirow{3}{1.4cm}{Traffic Density}  & high & $9.64$ & \cellcolor{lightergray} & \cellcolor{lightergray} & 7.46 & 3.59 & 3.59\\
		& average & $1.6$ & \cellcolor{lightergray} & \cellcolor{lightergray} & 0.29  & 0.83 & 0.83 \\
		& \cellcolor{lightergray}low & \cellcolor{lightergray}$1.00$ & \cellcolor{lightergray}\multirow{-3}{*}{$4.64$} & \cellcolor{lightergray}\multirow{-3}{*}{$4.64$} & 0.17 & 0.77 & 0.77 \\
		\midrule
		\multirow{2}{1.4cm}{Object Distance} & far & $5.36$ & \cellcolor{lightergray} & \cellcolor{lightergray} & 5.36& 2.31 & 2.31 \\
		& \cellcolor{lightergray}close & \cellcolor{lightergray}$1.00$ &\cellcolor{lightergray}\multirow{-2}{*}{$2.31$} & \cellcolor{lightergray}\multirow{-2}{*}{$2.31$} & 0.19 & 0.43 & 0.43 \\
		\bottomrule
	\end{tabular}
	\caption{Categorical and dichomotic evaluation of $\RCE$, $\RRW$ and $\IRRW$. Reference values are highlighted in gray.}
	\label{tbl:acerce_single}
\end{table}
\vspace{-0.8cm}

\subsubsection{Multiple interventions}
Besides single interventions, it is also possible to calculate multiple, combined interventions $P(Y|\Do{X_1=x_1,X_2=x_2,...})$ \cite{pearl2009causality}. This resembles the cut sets analysis in FTA, as it exposes cases where multiple TCs are necessary for a performance decrease of performance. Although an arbitrary number of interventions is possible, in the following we focus on pairwise interventions. Analogously to equation \eqref{eq:ace_rce} we calculate the $\RCE^{2}$ as:
{\footnotesize
\begin{equation*}
\RCE_C^2 = \frac{P(Y|\Do{X_1=x_1,X_2=x_2})}{P(Y|\Do{X_1=x_{1,ref},X_2=x_{2,ref}})}.
\end{equation*}}
\begin{figure}[htb!]
	\centering
	\includegraphics[width=0.75\textwidth]{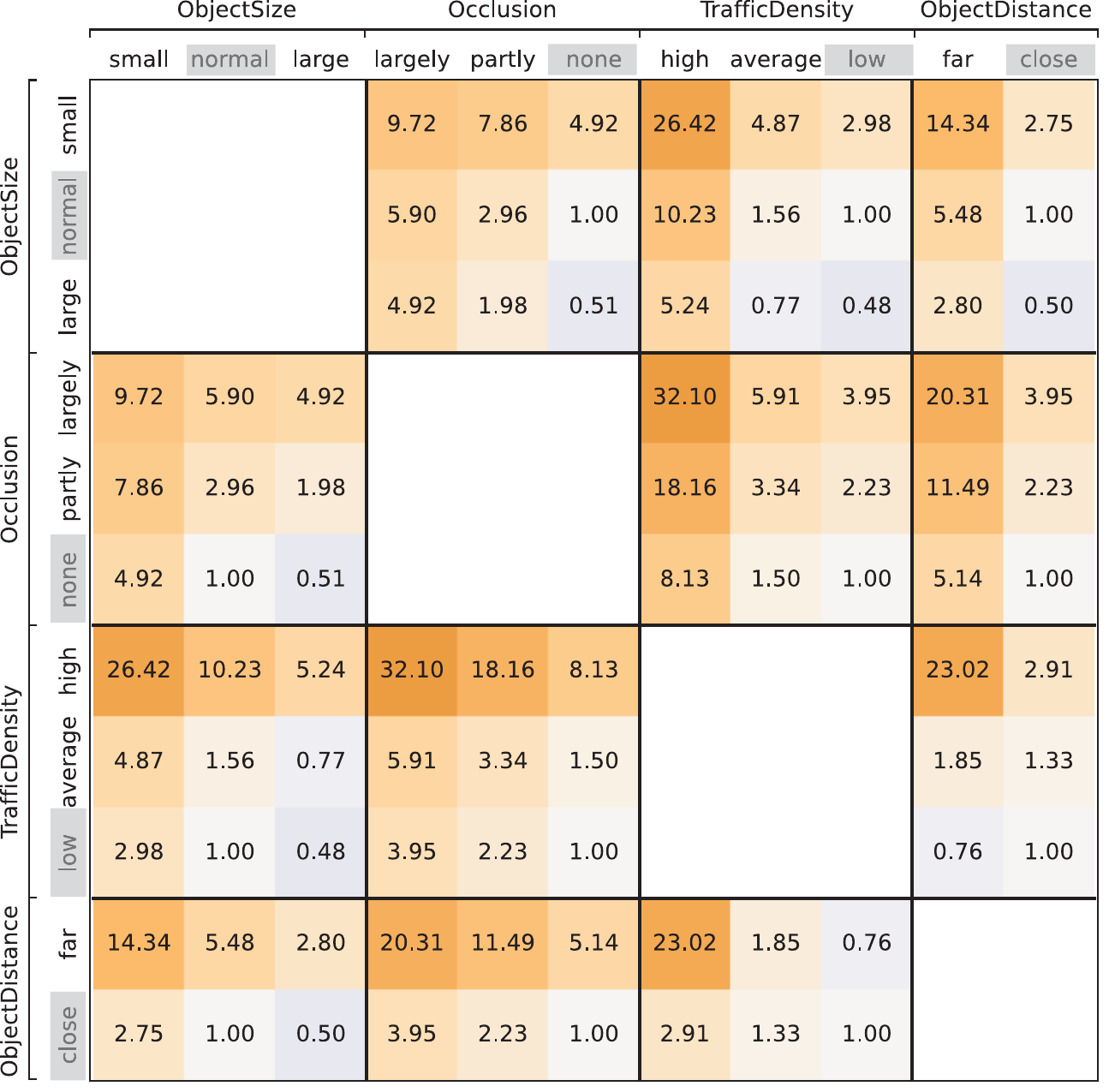}
	\caption{$\RCE_C^{2}$  for pairwise interventions on TCs with the grayed states used as reference. Each intervention combination is given by a row and column pair.}
	\label{fig:pairwise_intervention}
\end{figure}
\autoref{fig:pairwise_intervention} shows the $RCE^{2}$ for all pairwise combinations of TCs in our perception example. 
Notably, TCs with a high impact from single intervention are also pronounced in the pairwise interventions. 
This is not true in general, as a positive and negative causal impact from two nodes may cancel each other. 
The pairwise intervention (Occlusion=largely, TrafficDensity=high) exhibits the highest $\RCE^{2} \approx 32.1$, primarily due to the Fusion node, whose CPT resembles that of an AND-gate. 
Interventions that influence both causal paths to an AND-gate typically result in a high causal impact  
-- see also the combinations (TrafficDensity, ObjectSize) or (Occlusion, ObjectDistance).
In contrast, pairwise combinations located in just a single incoming path to the Fusion node are ranked relatively low, as the AND-characteristic suppresses the causal impact. 
We conclude that regarding pairwise inventions on TCs, the Fusion node is the most critical component -- as expected from a majority voting pattern. 
Therefore, improving on the Fusion node, e.g., the underlying algorithm, leads to a substantial FN rate reduction. 
Other safety measures should focus on individual contributors, i.e., (TrafficDensity=high, Occlusion=largely) and (TrafficDensity=high, ObjectSize=small), by fortifying perception algorithms against these.

\FloatBarrier
\subsubsection{Path-specific Interventions}
\label{subsec:path_specific_interventions}

In the CBN approach, the graph is no longer restricted to a tree.
Thus, there may be multiple paths linking a variable of interest to the outcome.
E.g., the CBN of Figure~\ref{fig:use_case_models} contains two different paths connecting 'TrafficDensity' and 'Fusion', namely $\pi_1: \text{TrafficDensity}\rightarrow \text{Occlusion} \rightarrow \text{Sen1}\rightarrow \text{Fusion}$ and $\pi_2:  \text{TrafficDensity}\rightarrow \text{Sen2}\rightarrow \text{Fusion}$.
The contribution to the overall causal effects can differ along such paths.
Therefore, to design precise safety mechanisms, an examination of the effects along individual paths is needed.
To achieve this, we suggest \emph{path-specific effects} that limit the scope of causal effects to individual paths \cite{shpitser2013counterfactual}.

\begin{wraptable}[14]{R}{0.45\textwidth}
	\vspace{-0.5cm}
	\centering
	\small
	\begin{tabular}{lllll}
		\toprule
		& & $\APE$  & $\RPE$& $\frac{\APE}{\ACE}$\\
		Path & State & \tiny{$\left(\times 10^{-4}\right)$} & &\\
		\midrule
		\multirow{3}{*}{$\pi_1$} & high & $0.08$ & $1.19$ & $0.02$\\
		& average & $0.03$ & $1.07$ & $0.12$ \\
		& \cellcolor{lightergray}low & \cellcolor{lightergray}$0.00$ &\cellcolor{lightergray}$1.00$ &\cellcolor{lightergray} - \\
		\midrule
		\multirow{3}{*}{$\pi_2$} & high & $2.86$ & $8.13$ & $0.82$ \\
		& average & $0.20$ & $1.50$ &$0.82$ \\
		& \cellcolor{lightergray}low& \cellcolor{lightergray}$0.00$ & \cellcolor{lightergray}$1.00$ & \cellcolor{lightergray} -\\
		\bottomrule
	\end{tabular}
	\caption{Categorical evaluation of path-specific effects of TrafficDensity on Fusion=FN.}
	\label{tab:path-specific-effect-comparison}
\end{wraptable}
The main idea is to model two interventions for a variable $X$ at the same time. Set $X=x$ for the path(s) $\pi$ under investigation and $X=x_{ref}$ for the remaining paths, denoted by $\DoOp_{\pi}(X=(x,x_{ref}))$.
For example, to investigate the path-specific effect of TrafficDensity=high on Fusion via the path $\pi_1$,  the distribution of TrafficDensity is replaced by setting TrafficDensity=high as input for Occlusion and simultaneously to a comparative value, such as $\neg$high or low,  as input for Sen2.
As in \autoref{subsec:safety_metric}, the comparative value can refer to the value's negation (dichotomic analysis) or to a reference value (categorical analysis).

Let us remark that the analysis of path-specific effects is a counterfactual query.
In general, the path-specific effect $\DoOp_{\pi}(X=(x,x_{ref}))$ on a variable $Y$ via a set of paths $\pi$ can be calculated form observational data if the causal effect $P(Y|\Do{X=x})$ is identifiable and the value assignment of $X$ is unambiguous. 
For DAGs without latent confounding the latter condition holds if  $\pi$ does not contain any causal paths from $X$ to $Y$ which start with the same arrow as a causal path from $X$ to $Y$ that is not in $\pi$, cf. \cite[Theorem 5]{avin2005identifiability}.
Figure \ref{fig:tornadochart_path} visualizes the path-specific effects of TrafficDensity on Fusion via different paths. 
The tornado chart shows the path-specific effects for $\pi = \{ \pi_1, \pi_2\}$ -- equivalent to the overall causal effect -- and then for $\pi_1$ and $\pi_2$ on their own.
The comparison of these path-specific effects indicates that almost the entire causal effect is transported via $\pi_2$.
For a more detailed analysis of path specific effects we introduce the following metrics
{\footnotesize
\begin{align*}
	\APE &= 	P(Y|\DoOp_{\pi}(X=(x, x_{ref})))- P(Y|\Do{X=x_{ref}}),\\
	\RPE &=  \frac{P(Y|\DoOp_{\pi}(X=(x,x_{ref})))}{P(Y|\Do{X=x_{ref}})},\\
	\frac{\APE}{\ACE}&=\frac{P(Y|\DoOp_{\pi}(X=(x,x_{ref})))- P(Y|\Do{X=x_{ref}})}{ P(Y|\Do{X=x}) - P(Y|\Do{X=x_{ref}})},
\end{align*}}
whose evaluation for $\pi_1$ and $\pi_2$ is given by \autoref{tab:path-specific-effect-comparison}.
The average and relative path-specific effects $\APE$ and $\RPE$ are defined analogously to $\ACE$ and $\RCE$, cf.~Shpitser et al.\footnote{The average path-specific causal effect is called 'effect along paths in $\pi$' \cite{shpitser2013counterfactual}.}, comparing an intervention to a comparative value for all paths.
In addition, the ratio of $\APE$ by $\ACE$ provides a comparison of the impacts via the investigated paths against via the whole model.
To interpret these metrics a comparison of the different paths is required. 
The values estimated for the example of \autoref{fig:use_case_models} are given in \autoref{tab:path-specific-effect-comparison}.

	\section{Conclusion and Future Work}
	\label{sec:conclusion}

In this work, we considered CBNs for the safety analysis of safety-critical complex systems. 
CBNs provide a promising alternative to FTA, particularly when dealing with complex dependencies. FTA is not suited to grasp the fault and failure propagation in such systems. Hence, CBNs become necessary to model and analyze causal relations to ensure SOTIF. The key advantage is the combined approach of systematically addressing uncertainties using data as well as expert knowledge.
To match FTA's quantification potential, we propose several causal importance metrics relying on causal inference.
To account for the complexity of CBNs we considered path-specific causal effects.
Finally, we evaluated the importance measures on an example perception system in the context of automated driving.

There are two main directions for future work. 
Firstly, the approach needs to be validated using real data coming from an actual complex system. 
As an intermediate step synthetic data from a simulation can be helpful. Secondly, when such data are available, causal learning (causal discovery) techniques can be integrated in the approach to obtain or verify parts of the causal graph.

%
	\bibliographystyle{splncs04}
	\bibliography{literature}

\begin{thebibliography}{10}
\providecommand{\url}[1]{\texttt{#1}}
\providecommand{\urlprefix}{URL }
\providecommand{\doi}[1]{https://doi.org/#1}

\bibitem{adee2021systematic}
Adee, A., Gansch, R., Liggesmeyer, P.: {Systematic Modeling Approach for
  Environmental Perception Limitations in Automated Driving}. In: 17th European
  Dependable Computing Conference. pp. 103--110 (2021)

\bibitem{adee2021discovery}
Adee, A., Gansch, R., Liggesmeyer, P., Glaeser, C., Drews, F.: {Discovery of
  Perception Performance Limiting Triggering Conditions in Automated Driving}.
  In: 5th International Conference on System Reliability and Safety. pp.
  248--257 (2021)

\bibitem{avin2005identifiability}
Avin, C., Shpitser, I., Pearl, J.: {Identifiability of Path-Specific Effects}.
  In: Proceedings of the 19th International Joint Conference on Artificial
  Intelligence. p. 357–363. IJCAI'05, San Francisco, CA, USA (2005)

\bibitem{avizienis2004basic}
Avizienis, A., Laprie, J.C., Randell, B., Landwehr, C.: Basic concepts and
  taxonomy of dependable and secure computing. IEEE transactions on dependable
  and secure computing  \textbf{1}(1),  11--33 (2004)

\bibitem{birnbaum1968importance}
Birnbaum, Z.W.: On the importance of different components in a multicomponent
  system. Tech. rep., University of Washington, Seattle (1968)

\bibitem{damm2018perspectives}
Damm, W., Fr{\"a}nzle, M., Gerwinn, S., Kr{\"o}ger, P.: {Perspectives on the
  Validation and Verification of Machine Learning Systems in the Context of
  Highly Automated Vehicles.} In: AAAI Spring Symposia (2018)

\bibitem{ducamp2020agrum}
Ducamp, G., Gonzales, C., Wuillemin, P.H.: {aGrUM/pyAgrum : a toolbox to build
  models and algorithms for Probabilistic Graphical Models in Python}. In:
  Proceedings of the 10th International Conference on Probabilistic Graphical
  Models. vol.~138, pp. 609--612 (2020)

\bibitem{deletang2021causal}
Déletang, G., Grau-Moya, J., Martic, M., Genewein, T., McGrath, T., Mikulik,
  V., Kunesch, M., Legg, S., Ortega, P.A.: {Causal Analysis of Agent Behavior
  for AI Safety} (2021)

\bibitem{espiritu2007component}
Espiritu, J.F., Coit, D.W., Prakash, U.: Component criticality importance
  measures for the power industry. Electric Power Systems Research
  \textbf{77}(5) (2007)

\bibitem{fenton2018risk}
Fenton, N., Neil, M.: {Risk Assessment and Decision Analysis with Bayesian
  Networks}. CRC Press, 2 edn. (2018)

\bibitem{iso21448}
{International Organization for Standardization (ISO)}: {ISO 21448: Road
  vehicles -- Safety of the intended functionality} ({2022})

\bibitem{issa2022use}
Issa~Mattos, D., Liu, Y.: {On the Use of Causal Graphical Models for Designing
  Experiments in the Automotive Domain}. In: Proceedings of the International
  Conference on Evaluation and Assessment in Software Engineering 2022. p.
  264–265. EASE '22, New York, NY, USA (2022)

\bibitem{jesenski2019generation}
Jesenski, S., Stellet, J.E., Schiegg, F., Z{\"o}llner, J.M.: Generation of
  scenes in intersections for the validation of highly automated driving
  functions. In: 2019 IEEE Intelligent Vehicles Symposium (IV). pp. 502--509.
  IEEE (2019)

\bibitem{jiang2024generation}
Jiang, Z., Liu, J., Sun, P., Sang, M., Li, H., Pan, Y.: Generation of risky
  scenarios for testing automated driving visual perception based on causal
  analysis. IEEE Transactions on Intelligent Transportation Systems  (2024)

\bibitem{koopmann2025grasping}
Koopmann, T., Putze, L., Westhofen, L., Gansch, R., Adee, A., Neurohr, C.:
  {Grasping Causality for the Explanation of Criticality for Automated
  Driving}. IEEE Access  \textbf{13},  54739--54756 (2025)

\bibitem{kramer2020identification}
Kramer, B., Neurohr, C., B{\"u}ker, M., B{\"o}de, E., Fr{\"a}nzle, M., Damm,
  W.: {Identification and Quantification of Hazardous Scenarios for Automated
  Driving}. In: Model-Based Safety and Assessment. pp. 163--178 (2020)

\bibitem{maier2024testing}
Maier, R., Grabinger, L., Urlhart, D., Mottok, J.: {Causal Models to Support
  Scenario-Based Testing of ADAS}. IEEE Transactions on Intelligent
  Transportation Systems  \textbf{25}(2),  1815--1831 (2024)

\bibitem{neurohr2021criticality}
{Neurohr}, C., {Westhofen}, L., {Butz}, M., {Bollmann}, M.H., {Eberle}, U.,
  {Galbas}, R.: {Criticality Analysis for the Verification and Validation of
  Automated Vehicles}. IEEE Access  \textbf{9},  18016--18041 (2021)

\bibitem{niu2023causal}
Niu, Y., Fan, Y., Gao, Y., Li, Y.: A causal inference method for improving the
  design and interpretation of safety research. Safety Science  \textbf{161},
  106082 (2023)

\bibitem{nyberg2013failure}
Nyberg, M.: {Failure propagation modeling for safety analysis using causal
  Bayesian networks}. In: Conference on Control and Fault-Tolerant Systems. pp.
  91--97 (2013)

\bibitem{pearl2009causality}
Pearl, J.: Causality. Cambridge University Press, 2 edn. (2009)

\bibitem{peters2017}
Peters, J., Janzing, D., Schölkopf, B.: {Elements of Causal Inference:
  Foundations and Learning Algorithms}. The MIT Press (2017)

\bibitem{qiu2021parameter}
Qiu, M., Kryda, M., Bock, F., Antesberger, T., Straub, D., German, R.:
  Parameter tuning for a markov-based multi-sensor system. In: 47th Euromicro
  Conference on Software Engineering and Advanced Applications. pp. 351--356.
  IEEE (2021)

\bibitem{ruijters2015}
Ruijters, E., Stoelinga, M.: {Fault tree analysis: A survey of the
  state-of-the-art in modeling, analysis and tools}. Computer Science Review
  \textbf{15-16},  29--62

\bibitem{shpitser2013counterfactual}
Shpitser, I.: {Counterfactual Graphical Models for Longitudinal Mediation
  Analysis With Unobserved Confounding}. Cognitive Science  \textbf{37}(6),
  1011--1035 (2013)

\bibitem{skjong2001expert}
Skjong, R., Wentworth, B.H.: Expert judgment and risk perception. In: ISOPE
  International Ocean and Polar Engineering Conference. pp. ISOPE--I. ISOPE
  (2001)

\bibitem{thomas2023toward}
Thomas, S., Groth, K.M.: Toward a hybrid causal framework for autonomous
  vehicle safety analysis. Proceedings of the Institution of Mechanical
  Engineers, Part O: Journal of Risk and Reliability  \textbf{237}(2),
  367--388 (2023)

\bibitem{werling2025safety}
Werling, M., Faller, R., Betz, W., Straub, D.: Safety integrity framework for
  automated driving. arXiv preprint arXiv:2503.20544  (2025)

\bibitem{zeller2022component}
Zeller, M.: {Component Fault and Deficiency Tree (CFDT): Combining Functional
  Safety and SOTIF Analysis}. In: {Model-Based Safety and Assessment: 8th
  International Symposium, IMBSA, Munich, Germany}. pp. 146--152. Springer
  (2022)

\end{thebibliography}
	
	\appendix
	\newpage
	\section{Appendix - Conditional Probability Tables}
	\label{sec:appendix}


\setlength{\tabcolsep}{3pt}

\begin{table}[h]
	\caption{Conditional probability tables for the confounding example of Figure~\ref{fig:confounding_example}.}
	\tiny
	\centering
	\begin{tabular}{ccccc}
		\toprule
		& & \multicolumn{3}{c}{Luminance} \\
		\multicolumn{2}{c}{Weather} & low & medium & high \\ \midrule
		0.6 & sun & 0.05 & 0.4 & 0.55 \\
		0.3 & rain & 0.2 & 0.6 & 0.2 \\
		0.1 & snow & 0.6 & 0.3 & 0.1 \\
		\bottomrule
		\toprule
			& & \multicolumn{3}{c}{Brightness (correlation)}\\
		&	Luminance & low & medium & high \\ \midrule
		&	low & 0.9 & 0.1 & 0\\
	&		high & 0 & 0 & 1	\\
			\bottomrule	
				\toprule
		&	& \multicolumn{3}{c}{Brightness (causal)}\\
		&	Luminance & low & medium & high \\ \midrule
		&	low & 0.9 & 0.1 & 0\\
		&	medium & 0 & 1 & 0\\
		&	high & 0 & 0.4 & 0.6 \\
			\bottomrule		
	\end{tabular}
	\qquad
	\begin{tabular}{C{2cm}ccc}
		\toprule
		& & \multicolumn{2}{c}{Perception}\\
		Luminance (Brightness) & Weather & FN & TP \\
		\midrule
		& sun & 0.04 & 0.96\\
		low & rain & 0.075 & 0.925\\
		& snow & 0.11 & 0.89\\ 	\midrule
		& sun & 0.035 & 0.965\\
		medium & rain & 0.07 & 0.93\\
		& snow & 0.09 & 0.91\\ 	\midrule
		& sun & 0.04 & 0.0.96\\
		high & rain & 0.08 & 0.92\\
		& snow & 0.105 & 0.895\\
		\bottomrule
		\end{tabular}
	\label{tab:conditional_probabilities_confounding}
\end{table}

\begin{table}[h]
	\caption{Conditional probability tables for the domain nodes in \autoref{sec:example}.}
	\centering
	\tiny
	\begin{tabular}{cccc}
		\toprule
		\multicolumn{2}{c}{ObjectSize} & \multicolumn{2}{c}{TrafficDensity} \\ \midrule
		small & 0.2 & high & 0.4 \\
		normal & 0.4 & average & 0.3  \\
		large & 0.4 & low & 0.3 \\
		\bottomrule
		& & & \\
		\toprule
		\multicolumn{2}{c}{ObjectDistance} & \multicolumn{2}{c}{Occlusion (FTA)}\\ \midrule
	far & 0.3 & largely& 0.1\\
	close & 0.7 & partly&0.45\\
	&&none&0.45\\
		\bottomrule
	\end{tabular}
	\quad
	\begin{tabular}{ccccc}
		\toprule
		& & \multicolumn{3}{c}{Occlusion (CBN)}\\
		ObjectSize & TrafficDensity & largely & partly & none \\
		\midrule
		& high & 0.27 & 0.4 & 0.33 \\
	  small	& average & 0.15 & 0.6 & 0.25 \\
		& low & 0.05 & 0.55 & 0.4 \\ \midrule
		 & high & 0.2 & 0.45 & 0.35 \\
	 normal	& average & 0.1 & 0.45 & 0.45 \\
		& low & 0.1 & 0.4 & 0.5 \\ \midrule 
	 & high & 0.05 & 0.5 & 0.45 \\
	large	& average & 0.01 & 0.42 & 0.57 \\
		& low & 0.01 & 0.3715 & 0.6185 \\
		\bottomrule
	\end{tabular}
	\label{tab:conditional_probabilities_use_case_domain_nodes}
\end{table}

\vspace{-1.4cm}
\begin{table}[b!]
	\caption{Conditional probability tables for the system nodes in \autoref{sec:example}.}
	\centering
	\tiny
	\begin{tabular}{cccc}
		\toprule
		& & \multicolumn{2}{c}{Sen1}\\
		ObjectSize & Occlusion & FN & TP \\ \midrule
		 & largely & 0.0495 & 0.9505\\
		small & partly & 0.04 & 0.96\\
		& none & 0.025 & 0.975\\ \midrule
		 & largely & 0.03 & 0.97\\
		normal & partly & 0.015 & 0.985\\
		& none & 0.005 & 0.995\\ \midrule
		 & largely & 0.025 & 0.975\\
		large & partly & 0.01 & 0.99\\
		& none & 0.0025 & 0.9975\\
		\bottomrule
		\toprule
		& & \multicolumn{2}{c}{Fusion}\\
		Sen1 & Sen2 & FN & TP \\ \midrule
		\multirow{2}{*}{FN} & FN & 0.95 & 0.05\\
		& TP & 0.0001 & 0.9999\\ \midrule
		\multirow{2}{*}{TP} & FN & 0.0001 & 0.9999\\
		& TP & 0 & 1\\
		\bottomrule
	\end{tabular}
	\quad
	\begin{tabular}{C{1.3cm}C{1.3cm}cc}
		\toprule
		& & \multicolumn{2}{c}{Sen2}\\
		TrafficDensity & ObjectDistance & FN & TP \\ \midrule
		\multirow{2}{*}{high} & far & 0.064 & 0.936\\
		& close & 0.008 & 0.992\\ \midrule
		\multirow{2}{*}{average} & far & 0.0056 & 0.9944\\
		& close & 0.004& 0.996\\ \midrule
		\multirow{2}{*}{low} & far & 0.0024 & 0.9976\\
		& close & 0.0032 & 0.9968\\
		\bottomrule
	\end{tabular}
	\label{tab:conditional_probabilities_use_case_system_nodes}
\end{table}

\end{document}